\pdfoutput=1

\documentclass[11pt]{article}

\usepackage[preprint]{acl}

\usepackage{times}
\usepackage{latexsym}
\usepackage{graphicx}

\usepackage[T1]{fontenc}

\usepackage{amsmath}
\usepackage{multirow}
\usepackage{multicol}
\usepackage{mathabx}
\usepackage{algorithm,algpseudocode}

\usepackage{mathtools}
\usepackage{microtype}

\title{TROPE: TRaining-Free Object-Part Enhancement for Seamlessly Improving Fine-Grained Zero-Shot Image Captioning}

\author{Joshua Feinglass \textnormal{and} Yezhou Yang\\
  Arizona State University\\
  \texttt{\{joshua.feinglass,yz.yang\}@asu.edu} 
  }

\begin{document}
\maketitle
\begin{abstract}
Zero-shot inference, where pre-trained models perform tasks without specific training data, is an exciting emergent ability of large models like CLIP. Although there has been considerable exploration into enhancing zero-shot abilities in image captioning (IC) for popular datasets such as MSCOCO and Flickr8k, these approaches fall short with fine-grained datasets like CUB, FLO, UCM-Captions, and Sydney-Captions. These datasets require captions to discern between visually and semantically similar classes, focusing on detailed object parts and their attributes. To overcome this challenge, we introduce TRaining-Free Object-Part Enhancement (TROPE). TROPE enriches a base caption with additional object-part details using object detector proposals and Natural Language Processing techniques. It complements rather than alters the base caption, allowing seamless integration with other captioning methods and offering users enhanced flexibility. Our evaluations show that TROPE consistently boosts performance across all tested zero-shot IC approaches and achieves state-of-the-art results on fine-grained IC datasets\footnote{TROPE source codes and data:  \url{https://github.com/JoshuaFeinglass/TROPE}.}.
\end{abstract}
\begin{figure}
\centering
\def\svgwidth{\columnwidth}
\includegraphics[width=\columnwidth]{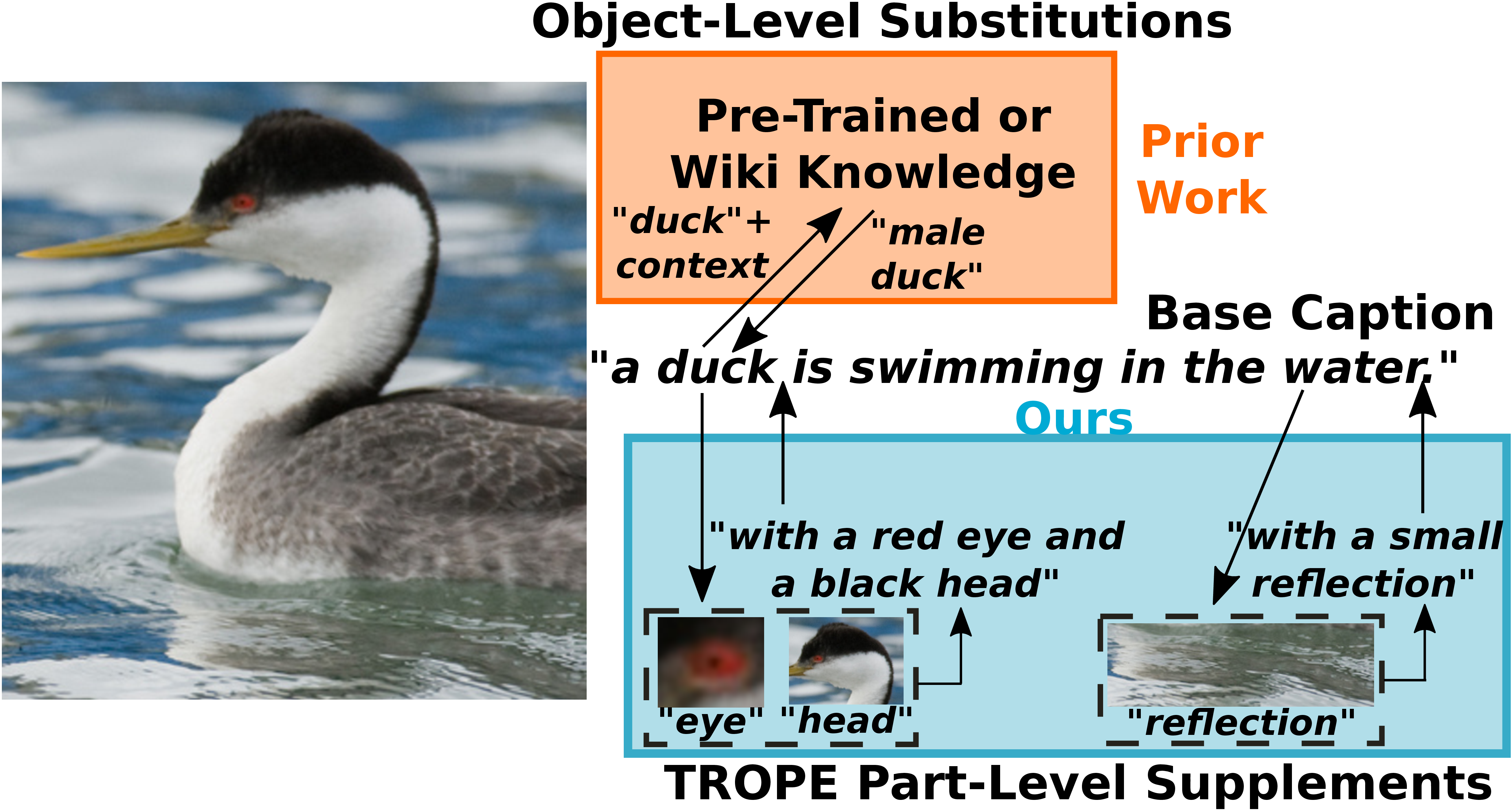}
\caption{An example differentiating TROPE from prior work in image caption enhancement, which substitute existing words in the sentence with more contextually appropriate alternatives. TROPE instead inserts supplemental information after key objects by mapping nouns to a region of the image and constructing semantic part proposals based on object parts and attributes found within this region.}
\label{fig:main}
\end{figure}
\section{Introduction}
Object parts and their attributes have been shown to play a critical role in distinguishing between classes in tasks like fine-grained classification \citep{partobject2024,domain-invariant2023,hawk2024}. Despite their importance, previous works in image captioning (IC) have instead focused primarily on objects, their attributes, and their interactions, as seen in by common utilized semantic structures like scene graphs \citep{memcap2020,sggcap2022,capsgg2020}. This focus is partly because IC is often applied to general domain datasets like MSCOCO \citep{coco2014} and Flickr8k \citep{flickr2013}, where images typically contain various object classes, and captions provide a high-level scene description. 
\par
Existing training-free Zero-Shot methods, such as ZeroCap \citep{zerocap2022} and ConZIC \citep{conzic2023}, enhance captions by substituting words in a base caption with more a contextually appropriate ones, utilizing scores from CLIP \citep{clip2021} and a Large Language Model (LLM). However, since these models are pre-trained on general domain data, the resulting captions often lack fine-grained detail. This is especially problematic for fine-grained datasets, such as bird species in CUB \citep{cub2010}, flower species in FLO \citep{cub2010}, or aerial scenes in UCM-Captions \citep{ucm2010} and Sydney-Captions \citep{ucm2010}, which require distinguishing between visually and semantically similar classes. \citet{gap2023} showed that that visual and semantic feature representations from these fine-grained datasets differ significantly from those of general domain datasets, leading to poor performance in domain generalization benchmarks due to task misalignment \citep{misalignment2024}. \par
We conjecture that effective zero-shot IC in fine-grained contexts necessitates robust primitives that are consistent across both the training and test domains. Following this line of reasoning, we proposed TRaining-free Object-Part Enhancement (TROPE), which adapts pre-trained models to fine-grained datasets by supplementing captions with object part information. TROPE effectively augments the base captions of existing zero-shot IC methods with fine-grained details as shown in Figure \ref{fig:main}. Our evaluations demonstrate that adding information from object part semantic proposals consistently enhances IC performance across all tested methods, datasets, and metrics. Precision-recall curves indicate that TROPE significantly improves recall with a minimal impact on precision, particularly in datasets where there is substantial overlap between the object detector's vocabulary and the terms commonly used by human annotators. We also present examples of TROPE's application to an enterprise captioner, GPT4, and discuss two failure cases: one involving a lack of recognizable objects and another featuring redundant or incorrect part information.
\par
To further explore the bias of general domain datasets, we conducted an analytical study on the frequency of terms in human-annotated and machine-generated texts across both general domain and fine-grained datasets. We found that semantic indicator words, such as "with", "has", and "have", which introduce object part descriptions, are much more common in fine-grained datasets. This finding underscores the strong relationship between the semantic structure of images and the captions used to describe them, reinforcing the need to adapt models trained on general domain datasets to fine-grained settings using techniques like TROPE.
\\
\noindent \textbf{Contributions:} 
Our work introduces the setting of fine-grained zero-shot captioning, extending zero-shot capabilities to four fine-grained captioning datasets. Our analyses reveal that existing zero-shot benchmarks cater predominantly to general domains and fail to meet the specific needs of fine-grained settings. We propose TROPE as a solution to enhance zero-shot captioning performance by incorporating detailed information from a pre-trained object detector, consistently enriching caption detail and improving performance across all methods, evaluation metrics, and datasets.
\section{Related Work}
The detection of objects and attributes, facilitated by large datasets of human-labeled regions \citep{vg2017,paco2023}, has historically been a cornerstone for various vision-language tasks \citep{vinvl2021}. Previous works have integrated this object and attribute information into training \citep{vinvl2021}, labels \citep{butd2018}, and text generation \citep{oscar2020}. TROPE builds on this foundation by extracting hierarchical object relationships to improve the detail of image captions.
\par
Enhancing the level of detail presented in image captions is a popular and multi-faceted topic. Entity-aware captioning seeks to replace generic nouns with context-specific entities from Wiki text based either on a base (template) caption \citep{entitytemplate2018,entitytemplate2019,entitytemplate2020} or optimized generation \citep{entityinform2020,entitytransform2020}. Similarly, stylized captioning is also performed by either modifying a base caption \citep{memcap2020} or optimized generation \citep{attractive2023}. Lastly, scene graphs have been used to either enhance a base caption \citep{memcap2020} or as an additional feature for optimized generation \citep{sggcap2022,capsgg2020}.
\par
Zero-shot captioning presents distinct challenges, as it operates without direct access to image-text pairs for training, relying instead on the intrinsic capabilities of pre-trained models like CLIP \citep{clip2021} and SimCTG \citep{simCTG2022}. Several works have tried to enhance zero-shot performance using training-free \citep{zeroshot2023,conzic2023} methods or text-only training \citep{plugging2022,decap2023, zerogen2023, textonly2022, textonly2023} strategies that assume access to target dataset captions. \citet{gap2023} introduced a domain generalization IC benchmark spanning general and fine-grained datasets, where a large gap in performance on general datasets and fine-grained datasets could be observed. TROPE aims to bridge the gap between general and fine-grained datasets by adding detailed object descriptions without requiring additional training data, improving the applicability and effectiveness of zero-shot captioning methods in more challenging environments.
\section{Preliminaries}
\subsection{Image Captioning Task}
The image captioning (IC) task involves an image captioner that takes an image as input and outputs a caption, typically a single sentence, that describes the image. The process begins with a vision module \textbf{V} that extracts features $w$ from the image as a pre-processing step. This is followed by a cross-modal understanding module \textbf{VL}, which integrates the pre-processed image information to generate the caption $y$ as shown
\begin{equation}
w\!=\!\textbf{V}(image), \ \ y\!=\!\textbf{VL}(w).
\end{equation}
\subsection{Object Detectors in Image Captioning}
n Vision-Language (VL) tasks such as IC, object detectors play a crucial role. These detectors are specialized to not only identify objects within an image but also provide detailed labels and attributes for these objects \citep{butd2018,vinvl2021}. The information about specific regions provided by these detectors is essential for many IC methods. For instance, Oscar \citep{oscar2020} model utilizes this detailed, region-specific data to facilitate cross-modal understanding when generating captions. Our work utilizes the object detector VinVL \citep{vinvl2021}, which provides bounding boxes $b_r$, regional features $\theta_r$, object labels $l^{o}_r$, and attribute labels $l^{a}_r$ (the most confident attribute label for an object) for all proposed regions of interest $r \in \mathcal{R}$ of an image 
\begin{equation}
    \{b_r,\theta_r,l_r^{o},l_r^{a}\}_{r \in \mathcal{R}} = \text{VinVL}(image).
\end{equation}
The integration of VinVL with the Oscar model is one of the approaches used in our work to generate base captions. We also select VinVL to serve as the source of object part information used by TROPE to enhance base captions because of its extremely large vocabulary of 1848, which encompasses objects present in all of the fine-grained datasets included in our benchmark.
\subsection{Measures of Object Proposal Similarity}
In assessing object proposals from VINVL, we focus on the spatial relationships and characteristics of the regions outlined by the bounding boxes. Operations such as intersection and union are used to evaluate the overlap between different bounding boxes, while the area of individual bounding boxes is calculated to assess their size. These measures help in determining the similarity and relevance of object proposals.
\begin{figure}
\centering
\def\svgwidth{\columnwidth}
\includegraphics[width=\columnwidth]{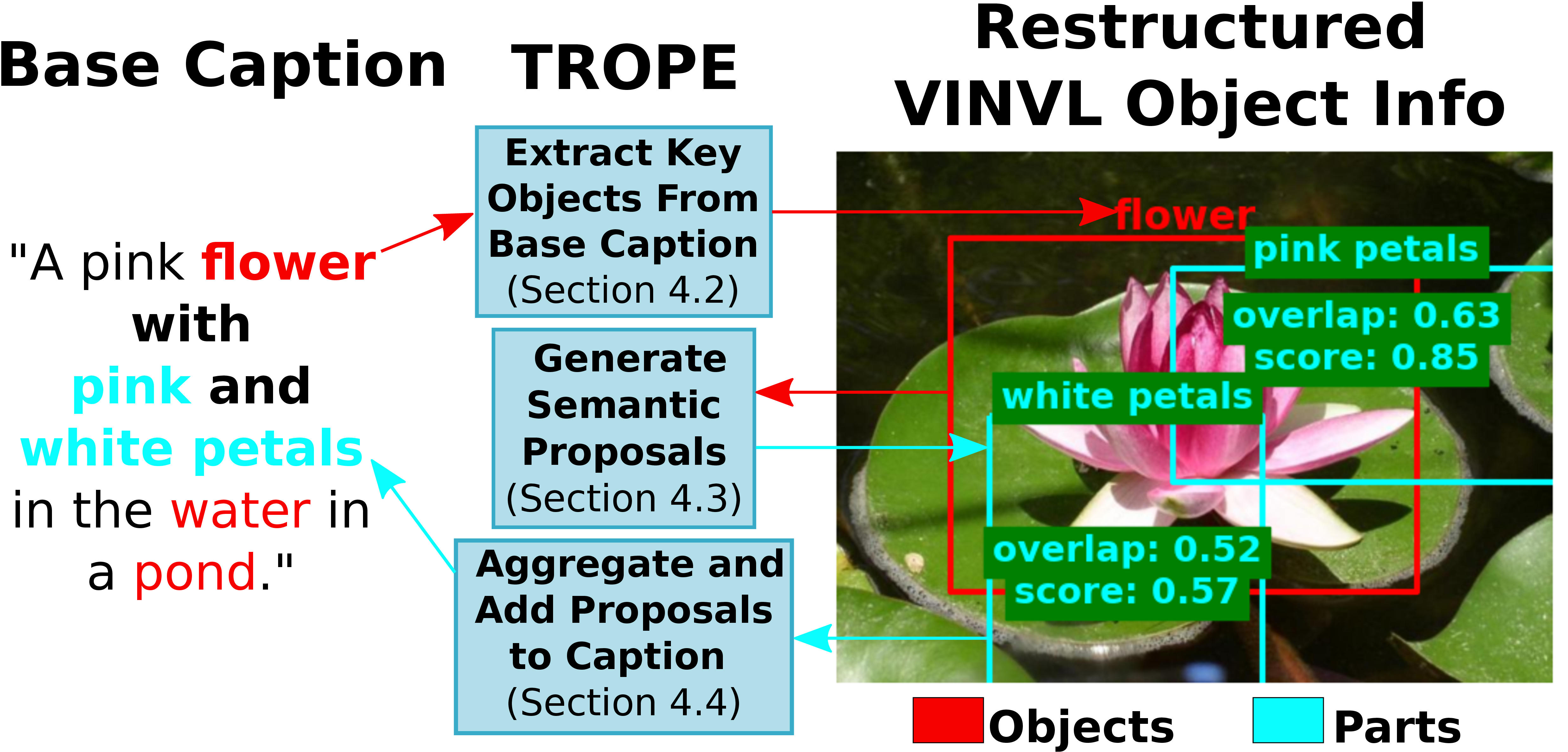}
\caption{A high-level visualization of the TROPE methodology expanded upon in Algorithm \ref{alg:trope}. Detailed descriptions of each TROPE function block can be found in their corresponding sections.}
\label{fig:method}
\end{figure}
\section{Structured Enhancement}
\subsection{Overview}
\label{sec:overview}
As depicted in Figure \ref{fig:method} and outlined in Algorithm \ref{alg:trope}, TROPE leverages raw object data from an object detector (VinVL) to enhance a base caption $y$ with supplemental text to form $y^+$. The additions to the base caption include semantic part proposals consisting of an article (if the object is singular), part attributes, and a part descriptor (e.g., "pink and white petals"). These proposals are associated with objects mentioned in the base caption (e.g., birds or flowers) and are integrated using punctuation and connective phrases such as ",", "and", "with", and "in addition to".
\subsection{Extract Key Objects from Base Caption ($h$)}
\label{sec:key}
Initially, key objects in the caption are identified. This involves extracting nouns and their corresponding caption indices from the base caption using tokenization and Parts-of-Speech (POS) tagging \citep{spacy2020}. For possessive phrases marked by "'s" or "of", only the possessing noun is considered, and the insertion index is set to the end of the phrase. As object detectors primarily recognize unigrams, compound nouns with space separation are not extracted. The identified nouns and indices are then matched with object labels and bounding boxes from VinVL, with plural nouns reverted to their singular form using Inflect \citep{inflect} to ensure consistent matching. The bounding box with the highest confidence is used for singular nouns, and the smallest box encompassing all relevant bounding boxes is used for plural nouns, resulting in bounding boxes $R^O$ and and corresponding caption indices $I^O$ for all key objects.
\subsection{Generate Semantic Proposals ($g$)}
\label{sec:gen}
The next process involves assigning smaller objects (indexed by $r$ in Algorithm \ref{alg:trope}) as parts of the larger key objects (indexed by $k$). Objects are sorted by the area of their bounding boxes, from smallest to largest, to prioritize smaller key objects and avoid assigning everything to large background objects like the "sky". Object proposals that overlap (defined as $area(b_k \cap b_r)/area(b_r)$ in Algorithm \ref{alg:trope}) significantly with a key object (exceeding a predefined threshold $T=0.5$ based on Table \ref{tab:ablationt}) are assigned as parts. If multiple parts share the same label and attribute, the label is pluralized using Inflect. Semantic part proposals $P$ are created by appending the attribute and part labels, prefacing with an article if the part is singular. Proposals in the list for an object $S_k$ are then ranked by adding overlap percentage and detector confidence (based on the ablation in Table \ref{tab:ablationsel}), determining their inclusion order.
\subsection{Aggregate and Add Semantic Proposals to Caption ($m$)}
\label{sec:enhance}
Redundant proposals, such as "white petals" and "pink petals", are organized coherently. Matching part labels with different attributes are combined using commas or "and" (e.g., "white and pink petals"). These semantic proposals are then inserted into the base caption at the identified object indices $I^O_k$. The number of proposals included per object is based on a user-defined parameter $N$. Proposals are introduced by "with", the most frequent semantic indicator from our word frequency study in Section \ref{sec:freq}. If an object's description already includes "with", "in addition to" is used to prepend the new proposals, ensuring a cohesive augmentation of the existing caption (e.g., "a flower with white and pink petals in addition to green leaves").
\begin{algorithm}
\begin{algorithmic}
\Require{$y$, $\{b_r,l^o_r,l^a_r\}_{r \in \mathcal{R}}$,$N$}
\Ensure{$y^+$} \Comment{Enhanced caption}
\State{$(R^O,I^O)=h(y)$} \Comment{Section \ref{sec:key}}
\For{$k \gets 1$ to $length(I^O)$}
\State{$S_k \gets \emptyset$}
\For{$r \in (\mathcal{R}-R^O)$}
\State{$S_k \gets \{\}$}
\If{$[area(b_k \cap b_r)/area(b_r)]>T$}

\State{$P \gets g(l^o_r,l^a_r)$} \Comment{Section \ref{sec:gen}}
 \State{$S_k$.append$(P)$}
\EndIf
\EndFor
\EndFor\\
Sort $I^O$ in reverse order \\
Apply the same reordering to $S_k$
\State{$y^+ \gets y$}
\For{$k \gets 1$ to $length(I^O)$}
\If{$S_k \neq \emptyset$}
\State{$y^+ \gets m(S_k,I^O_k,y^+,N)$} \Comment{Section \ref{sec:enhance}}
\EndIf
\EndFor
\end{algorithmic}
\caption{High-Level TROPE Pseudocode}
\label{alg:trope}
\end{algorithm}
\begin{table*}[t!]
\footnotesize
\centering
\scalebox{0.82}{
   \begin{tabular}{p{0.1cm} | l | c | c | c | c | c | c | c | c| c | c | c | c| c | c | c | c}
  	&  & \multicolumn{4}{c|}{\textbf{CUB}} & \multicolumn{4}{c|}{\textbf{FLO}}& \multicolumn{4}{c|}{\textbf{UCM}}& \multicolumn{4}{c}{\textbf{SC}}\\
   & \textbf{Method} & C & M & SP & SM &C & M & SP & SM &C & M & SP & SM &C & M & SP & SM \\
  	
  	\hline
  	
  	\multirow{4}{2em}{\rotatebox{90}{Domain Gen.}} 
     & Up-Down
  	& 3.70 & 7.99 & 14.56 & - & 9.04 & 8.07 & 12.38 & - & -& -& -& -& -& -& -& -\\
  	& AoANet
    	& 4.84 & 8.58 & 15.47 & - & 10.29 & 7.61 & 11.92 & - & -& -& -& -& -& -& -& -\\
  	&M$^2$Trans.
  	& 7.78 & 8.68 & 15.17 & - & 11.12 & 8.28 & 13.95 & - & -& -& -& -& -& -& -& -\\
    & EISNet
  	& 6.83 & 8.82 & 15.20 & - & 11.38 & 8.62 & 12.52 & - & -& -& -& -& -& -& -& -\\
  	& LSML
  	& 9.60 & 10.24 & 15.72 & - & 14.35 & 9.72 & \textbf{15.23} & - & -& -& -& -& -& -& -& -\\
  		\hline
  	  	\multirow{5}{2em}{\rotatebox{90}{Zero-Shot}} &
  	
  	ZeroCap & 0.33 & 4.75 & 0.25 & -1.14 & 0.47 & 5.12 & 0.34 & -1.13 & 0.59 & 5.39 & 1.84 & -1.13 & 0.32 & 4.31 & 0.60 & -1.14\\
  	& ConZIC & 10.30 & 10.17 & 2.02 & 0.41 & 15.07 & 11.32 & 3.19 & 0.78 & 7.54 & 6.51 & 2.88 & -0.27 & 12.33 & 7.62 & 3.48 & -0.05\\
   \cline{2-18}
    & \textbf{\rotatebox[origin=c]{180}{$\Lsh$}}1 part & 14.89 & 12.34 & 3.27 & 0.50 & 23.83 & 13.26 & 5.00 & 0.78 & \textbf{8.08} & 7.12 & 3.42 & -0.22 & \textbf{13.40} & \textbf{7.83} & 3.78 & \textbf{0.00}\\
    & \textbf{\rotatebox[origin=c]{180}{$\Lsh$}}5 parts & 7.21 & 14.31 & 5.80 & 0.58 & 16.21 & 14.06 & 5.60 & 0.81 & 6.64 & 7.16 & 3.36 & \textbf{-0.21} & 13.25 & 7.82 & 3.73 & \textbf{0.00}\\
    & \textbf{\rotatebox[origin=c]{180}{$\Lsh$}}10 parts & 5.73 & 14.10 & 6.56 & 0.55 & 16.16 & 14.05 & 5.62 & 0.81 & 6.64 & 7.15 & 3.35 & \textbf{-0.21} & 13.25 & 7.82 & 3.73 & \textbf{0.00}\\
    \cline{2-18}
   & Oscar & 29.63 & 15.52 & 6.40 & 0.26 & 41.56 & 15.15 & 7.67 & 0.32 & 7.95 & 7.31 & 4.85 & -0.83 & 5.12 & 6.19 & 2.62 & -0.82\\
   \cline{2-18}
    & \textbf{\rotatebox[origin=c]{180}{$\Lsh$}}1 part & \textbf{50.16} & 21.36 & 10.52 & 0.72 & \textbf{68.28} & 19.66 & 12.59 & 0.84 & 7.67 & 8.08 & 5.39 & -0.73 & 6.71 & 7.06 & 2.78 & -0.72\\
    & \textbf{\rotatebox[origin=c]{180}{$\Lsh$}}5 parts & 11.00 & \textbf{25.47} & 17.39 & \textbf{1.00} & 44.16 & \textbf{21.26} & 14.03 & \textbf{0.99} & 4.31 & \textbf{8.54} & \textbf{5.55} & -0.58 & 5.55 & 7.69 & 4.01 & -0.41\\
    & \textbf{\rotatebox[origin=c]{180}{$\Lsh$}}10 parts & 4.06 & 24.78 & \textbf{19.44} & 0.95 & 44.00 & 21.24 & 14.02 & \textbf{0.99} & 4.30 & 8.53 & 5.53 & -0.58 & 5.55 & 7.68 & \textbf{4.08} & -0.38\\

  \end{tabular}
}
 \caption{A fine-grained IC benchmark comparing the performance of domain generalization models from \citet{gap2023} including: Up-Down \citep{butd2018}, AoANet \citep{aoanet2019}, M$^2$Transformer \citep{meshed2020}, EISNet \citep{eisnet2020}, and LSML \citep{gap2023}, zero-shot IC methods including: ZeroCap \citep{zerocap2022}, ConZIC \citep{conzic2023}, and Oscar \citep{oscar2020}, and TROPE based enhancements of select zero-shot IC methods with varying numbers of semantic part proposals. Enhancements provided by TROPE are denoted with (\textbf{\rotatebox[origin=c]{180}{$\Lsh$}}).}
\label{tab:zero-shot}
\end{table*}
\section{Experiments}
To validate the effectiveness of TROPE for enhancing the detail of generated image captions, we conduct experiments on our proposed fine-grained IC benchmark in Section \ref{sec:benchmark}, where TROPE demonstrates consistent improvement in fine-grained image captioning performance for standard metrics. To further motivate the use of TROPE, we then explore the bias of general domain datasets in a word frequency study in Section \ref{sec:freq}.
\begin{figure*}[htb]
\centering
\def\svgwidth{\columnwidth}
\includegraphics[width=1.00\textwidth]{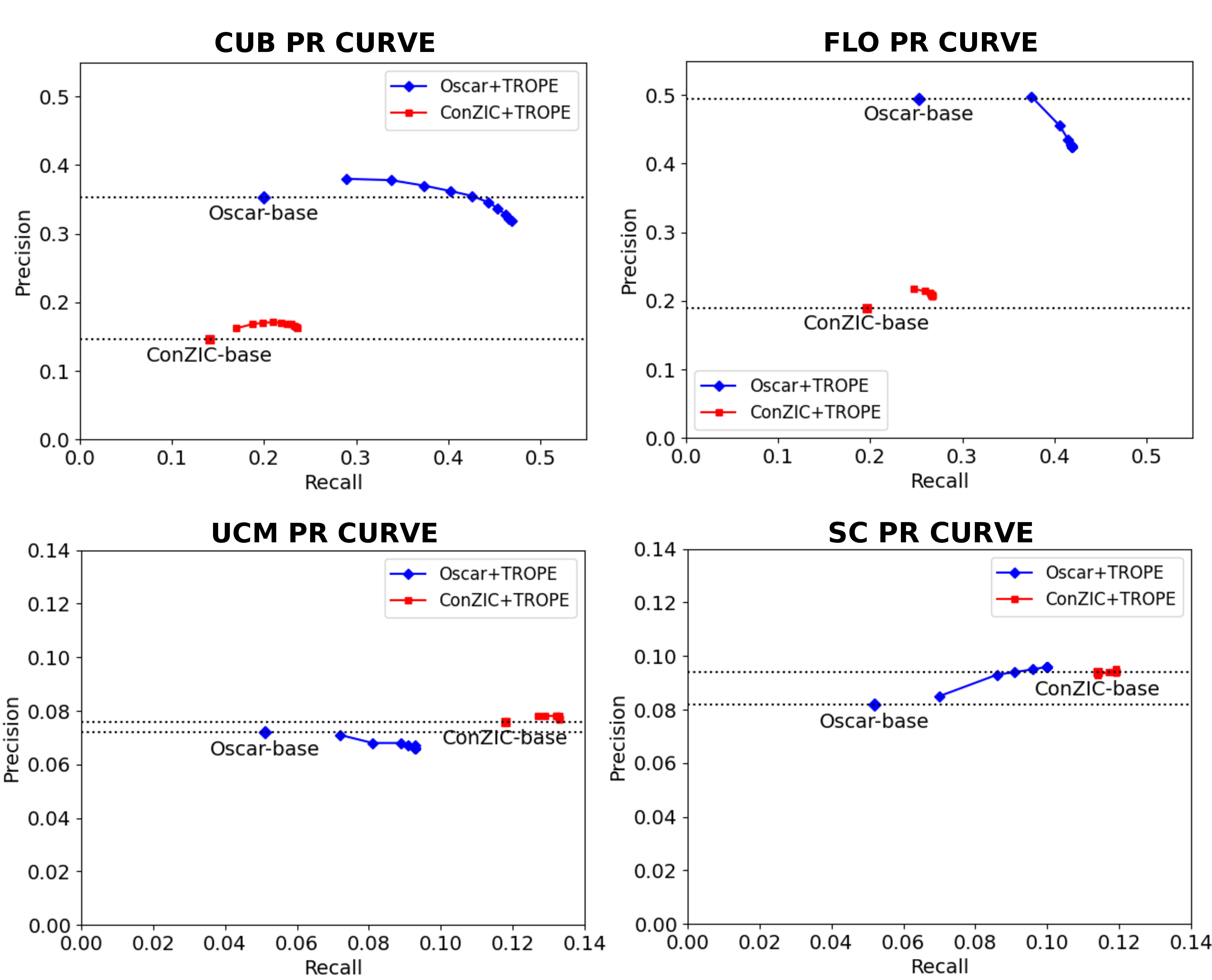}
\caption{Precision-recall curves generated by sweeping the number of semantic proposals added to the base caption from 1 to 10 for both the Oscar and ConZIC base captions. Horizontal lines represent the base caption precision performance.}
\label{fig:prcurve}
\end{figure*}
\begin{table}[htb]
\footnotesize
\centering
   \begin{tabular}{p{0.1cm}| c | c | c | c | c }
  	& & \multicolumn{2}{c|}{\textbf{CUB}} & \multicolumn{2}{c}{\textbf{FLO}} \\
   & \textbf{Criteria} & M & SP & M & SP \\
  	
  	\hline

    \multirow{3}{2em}{\rotatebox{90}{1 part}} & Score & \textbf{21.42} & 10.49 & 19.41 & 12.20 \\
  	& Overlap & 20.49 & 9.39 & 19.62 & \textbf{12.59}\\
    & Score+Overlap & 21.36 & \textbf{10.52} & \textbf{19.66} & \textbf{12.59}\\
    \hline
    \multirow{3}{2em}{\rotatebox{90}{5 parts}} & Score & 25.32 & 17.08 & 21.24 & 14.01 \\
    & Overlap & 24.97 & 16.53 & 21.21 & 14.00\\
    & Score+Overlap & \textbf{25.47} & \textbf{17.39} & \textbf{21.26} & \textbf{14.03}\\
  \end{tabular}
 \caption{An ablation study of the performance of different criteria for selecting proposals. Adding the object score (detector confidence) and overlap (from detector bounding boxes) yields the best captioning results.}
\label{tab:ablationsel}
\end{table}

\begin{table}[htb]
\footnotesize
\centering
   \begin{tabular}{p{0.1cm}| c | c | c | c | c }
  	& & \multicolumn{2}{c|}{\textbf{CUB}} & \multicolumn{2}{c}{\textbf{FLO}} \\
   & \textbf{Component} & M & SP & M & SP \\
  	
  	\hline

    \multirow{3}{2em}{\rotatebox{90}{1 part}} & Descriptor & 18.71 & 6.8 & 16.25 & 7.72 \\
  	& Part & 17.74 & 9.31 & 16.44 & 10.56\\
    & Both & \textbf{21.36} & \textbf{10.52} & \textbf{19.66} & \textbf{12.59}\\
    \hline
    \multirow{3}{2em}{\rotatebox{90}{5 parts}} & Descriptor & 22.82 & 7.03 & 18.37 & 8.11 \\
    & Part & 19.77 & 10.4 & 16.89 & 10.21\\
    & Both & \textbf{25.47} & \textbf{17.39} & \textbf{21.26} & \textbf{14.03}\\
  \end{tabular}
 \caption{An ablation study of the impact on performance when including only the descriptor or part component of the semantic proposal. The results suggest that METEOR is more sensitive to descriptors, while SPICE is more sensitive to object parts.}
\label{tab:ablationcomponent}
\end{table}
\subsection{Datasets}
\textbf{CUB} or the Caltech-UCSD Birds \citep{cub2010} dataset is a very popular benchmark for fine-grained classification and contains 200 classes of bird species (bobolink, cardinal, etc.). We use the 5,794 image test set with 10 captions for each image annotated by \citet{captions2016} for our benchmark.\\
\textbf{FLO} or the Oxford Flowers \citep{flo2008} is another popular benchmark for fine-grained classification and contains 102 classes of flower species (moon orchid, snapdragon, etc.). We use the 6,149 image test set with 10 captions for each image annotated by \citet{captions2016} for our benchmark.\\
\textbf{SC} or the Sydney Captions \citep{sydney2015} dataset consists of 7 land-use classes (residential, airport, etc.). We use the 58 image test set with 5 captions for each image annotated by \citet{satcaptions2016} for our benchmark.\\
\textbf{UCM} or the UC Merced Land Use \citep{ucm2010} dataset consists of 21 land-use classes (agricultural, harbor, etc.). We use the 210 image test set with 5 captions for each image annotated by \citet{satcaptions2016} for our benchmark.\\
\textbf{MSCOCO} or the Microsoft Common Objects in Common Context \citep{coco2014} dataset which is comprised of curated images containing 80 common object classes (like 'human' or 'truck') with 5 human annotated captions for each image.
\textbf{Flickr8k} \citep{flickr2013} is a popular dataset comprised of 8000 images with 5 human annotated captions for each image. The images were crawled from social media postings and like MS-COCO, are primarily common objects.
\begin{table}[htb]
\footnotesize
\centering
   \begin{tabular}{ c | c | c | c | c }
  	& \multicolumn{2}{c|}{\textbf{CUB}} & \multicolumn{2}{c}{\textbf{FLO}} \\
   \textbf{Threshold} & M & SP & M & SP \\
  	
  	\hline

    0.25 & 21.2 & 10.35 & \textbf{19.72} & 12.47 \\
  	0.50 & \textbf{21.36} & \textbf{10.52} & 19.66 & \textbf{12.59}\\
  	0.75 & 21.3 & 10.43 & 19.37 & \textbf{12.59}\\
  \end{tabular}
 \caption{An ablation study of the threshold $T$ used to assign parts to each object based on overlap for 1 part proposal. Although TROPE's performance does not seem to be strongly impacted by changes to $T$, a setting of $T=0.5$ exhibits the highest performance the majority of the time for the tested datasets and parameters.}
\label{tab:ablationt}
\end{table}
\subsection{Evaluation Metrics}
We utilize the four rule-based caption evaluation specific metrics which exhibit high agreement with human judgement across all commonly reported benchmarks: CIDEr (C) \citep{cider2015}, METEOR (M) \citep{meteor2005}, SPICE (SP) \citep{spice2016}, and SMURF (SM) \citep{smurf2021}. For all utilized metrics, a larger value indicators better performance with all metrics aside from SMURF varying within the range 0 to 1. SMURF is standardized to human performance, meaning a value of 0 is on par with human captions and negative or positive values indicate worse or better performance than humans, respectively. We exclude BLEU \citep{bleu2002} and ROUGE \citep{rouge2004} since they exhibit very poor agreement with human judgement in caption evaluation \citep{spice2016,smurf2021} and also do not consider metrics like CLIPScore \citep{clipscore2021} which are exclusively referenceless since they are likely to be sensitive to domain shift.
\subsection{Fine-Grained Captioning Benchmark}
\label{sec:benchmark}
Table \ref{tab:zero-shot} shows a comparison between base captioners enhanced using TROPE and relevant baselines. Zero-shot IC methods ZeroCap \citep{zerocap2022} and ConZIC \citep{conzic2023} utilizing pre-trained models CLIP \citep{clip2021}, BERT \citep{bert2019}, and GPT2 \citep{gpt2019} as well as the Oscar \citep{oscar2020} model utilizing VinVL \citep{vinvl2021} features are included since they are publicly available and achieve state-of-the-art results on MSCOCO \citep{coco2014} and Flickr8k \citep{flickr2013}. Domain generalization model results reported by \citet{gap2023} for CUB and FLO are also included in the benchmark because although they are not publicly available and train across four separate captioning sets, they still do not have access to target domain captions, making the setting zero-shot. The two most competitive and publicly available base captioners, Oscar and ConZIC, are selected for enhancement by TROPE. To aid in the design of TROPE, ablation studies utilizing the VinVL+Oscar pipeline shown in Tables \ref{tab:ablationsel}, \ref{tab:ablationcomponent}, and \ref{tab:ablationt} explore the impact of the proposal selection criteria, semantic components, and overlap threshold $T$ on captioning performance, respectively. To better show the trend of TROPE's performance for each additional proposal added to the base caption, we derive a precision and recall metrics from the SPARCS (SMURF's state-of-the-art semantic score) and use these metrics to generate precision and recall curves for each dataset shown in Figure \ref{fig:prcurve}. 
\begin{figure}[htb]
\centering
\def\svgwidth{\columnwidth}
\includegraphics[width=\columnwidth]{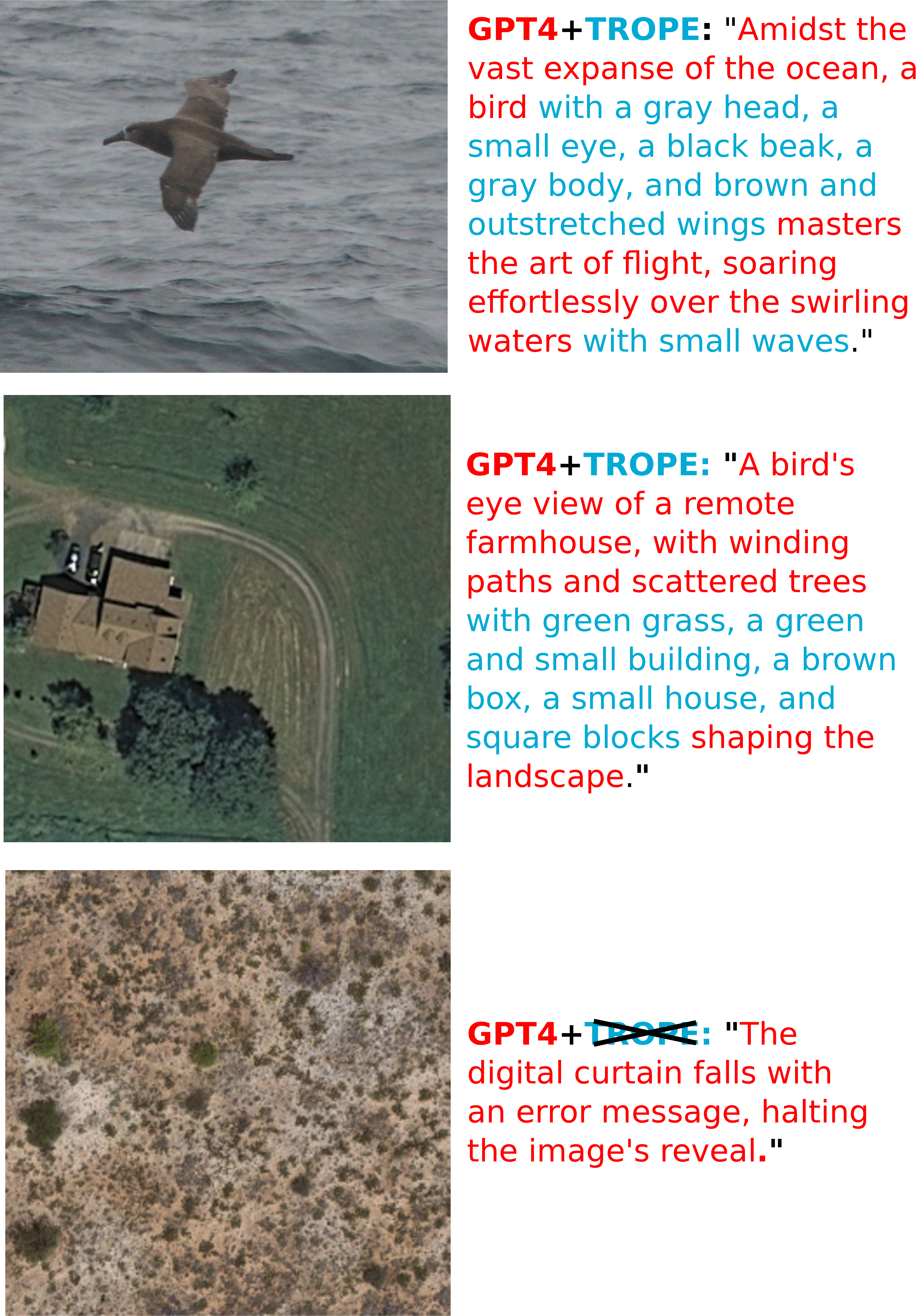}
\caption{Qualitative examples of TROPE applied to captions generated by GPT4 \citep{gpt2023} with $N=5$ semantic part proposals. Minor failures can be observed in the 2nd image caption with erroneous attributes like "green" house and redundant parts like "house" and "building". The 3rd caption is another failure case where no supplemental information from TROPE is added to the caption since the base caption contains no key objects detected by VinVL.}
\label{fig:gpt}
\end{figure}
\par
In general, the two state-of-the-art zero-shot IC methods achieved poor results compared to the rest of captioners. Zero-Cap in particular generated captions with little diversity and almost no relevance to the images. ConZIC performed significantly better, especially on aerial images when compared with Oscar, achieving the highest base model score for SMURF on UCM and the highest base model scores across all standard metrics on SC. Oscar achieved highly competitive results, especially on CUB and FLO, where it was the highest performing base model for all reported metrics except SMURF and SPICE. Inference for both ConZIC and Zero-Cap is extremely slow, taking more than a day to generate captions for the benchmark compared to the VinVL+Oscar pipeline which took a few hours. 
\par
Performance achieved by the Oscar and ConZIC increased significantly across all standard metrics, datasets, and tested models after adding 1 semantic part proposal. This can be attributed to a large jump in recall performance across all datasets in Figure \ref{fig:prcurve}, which then increases less significantly with each additional semantic part proposal added. Conversely, precision typically changes slightly with the first proposal, then decreases at an increasing rate with each additional proposal, with SC as a notable exception. These findings are discussed further in Section \ref{sec:discussion}. Although enterprise models like GPT4 \citep{gpt2023} are not included in the benchmark due to cost and rate limitations, we show 3 examples of TROPE integrated with GPT4 in Figure \ref{fig:gpt}, with 2 examples demonstrating improvements in caption detail and 1 example demonstrating a failure case with no change to the base caption.
\begin{figure}[t!]
\centering
\def\svgwidth{\columnwidth}
\includegraphics[width=\columnwidth]{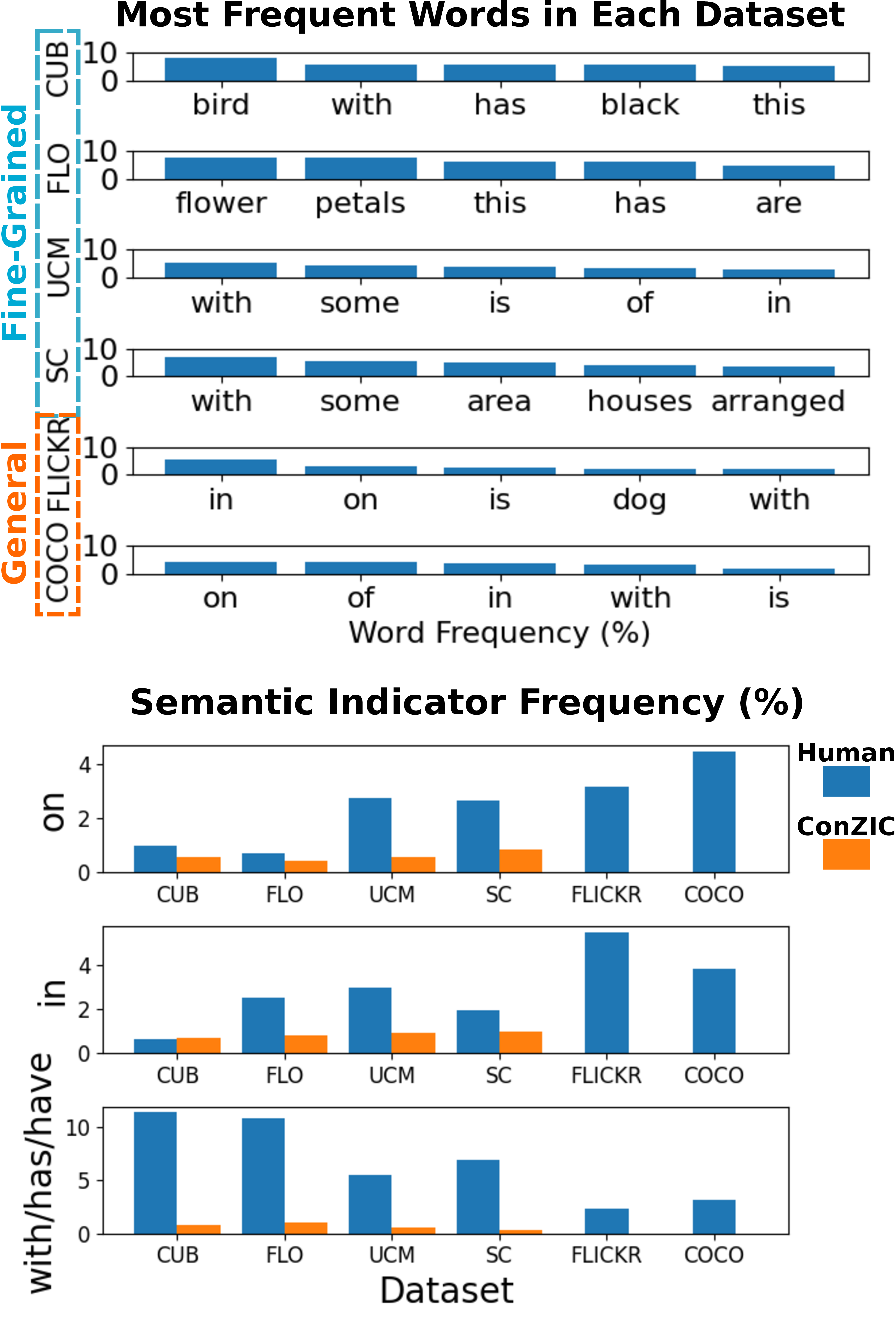}
\caption{Visualizations showcasing the unique characteristics of fine-grained datasets. The top plot shows the frequency of the 5 most common terms in our selected fine-grained and general domain datasets. The bottom plot shows the frequency of different semantic indicators across our selected datasets for both human annotations and available base captions from ConZIC.}
\label{fig:freq}
\end{figure}
\subsection{Word Frequency Study}
\label{sec:freq}
To explore the distinctive features of fine-grained captions, we analyzed the word frequency statistics of the training sets of our selected fine-grained image captioning datasets alongside general domain datasets such as MSCOCO and Flickr8k. The findings, illustrated in Figure \ref{fig:freq}, reveal distinct linguistic patterns between these dataset categories. Words that serve as semantic indicators of object-to-object interactions, such as 'on' and 'in', appear with greater frequency in general domain datasets. In contrast, words that indicate object-part descriptions, like 'with', 'has', and 'have', are more prevalent in fine-grained datasets.
\par
This variation in word usage underscores the unique requirements of fine-grained captioning, which often necessitates detailed descriptions of object parts and attributes. The state-of-the-art zero-shot IC captioning method, ConZIC, exhibits a notable deficiency in incorporating these semantic indicators for object-part descriptions, which likely contributes to its underperformance in fine-grained tasks. This observation supports our hypothesis that effective fine-grained captioning relies heavily on the precise depiction of object parts and attributes.
\par
Moreover, the word frequencies highlight that the granularity of fine-grained properties exists on a spectrum. Datasets like CUB and FLO, which typically feature a single salient object such as a "bird" or "flower", exhibit a high degree of specificity. Aerial datasets like UCM and SC, however, occupy a middle ground between general domain and fine-grained datasets. Although these datasets may include dominant objects like "airport" or "ocean", they lack the intense focus on singular objects characteristic of the most fine-grained datasets. This spectrum of granularity provides further context for tailoring image captioning approaches to suit the specific demands of different dataset types.
\section{Discussion}
\label{sec:discussion}
Based on our results and the apparent spectrum of fine-grained dataset characteristics, TROPE’s effectiveness appears widely applicable to numerous image datasets. However, its performance varies depending on each dataset's structure. A significant factor influencing TROPE's success is the alignment between the common terminology used by human captioners and the vocabulary of the object detector employed. For instance, while VinVL effectively covers common terms related to bird parts (e.g., head, tail, wing), flower parts (e.g., petal, leaf), and aerial views (e.g., airplane, airport), it lacks specialized terms frequently used in flower descriptions (e.g., stamen, pistil, veins). This gap is notable in our results: TROPE shows state-of-the-art performance on CUB, UCM, and SC, but somewhat underperforms in the FLO dataset compared to domain generalization techniques, particularly as additional part proposals are integrated, which dramatically affects precision.
\par
Furthermore, as the precision of object detectors improves, we anticipate that methods like TROPE will yield even greater improvements in image captioning performance. TROPE's strength lies in significantly boosting recall with minimal reductions in precision. In cases of poor detector performance, the typical outcome is no change to the base caption, whereas a mismatch between the detector’s vocabulary and human captions can lead to redundant or irrelevant descriptions, thereby decreasing precision.
\par
Our analysis also indicates that different caption evaluation metrics prioritize different aspects of TROPE’s semantic components (see Table \ref{tab:ablationcomponent}) and precision-recall performance curve. METEOR, SPICE, and SMURF achieve their highest scores with the incorporation of five additional parts per object, suggesting a preference for detailed content. Conversely, CIDEr peaks with just one additional part, likely because it penalizes excessive wordiness beyond the average reference caption length, which may not suit fine-grained captioning settings where detailed descriptions are crucial.
\par
Considering these insights, the optimal number of semantic part proposals to add to a base caption depends on the specific needs and goals of the research. For applications requiring high accuracy, such as assistive technologies, we recommend adding only a single proposal. Conversely, for purposes like training generative models or enhancing retrieval systems, incorporating multiple proposals may be beneficial as it enhances the discriminative information available, despite the potential for introducing irrelevant details. Researchers should select evaluation metrics that best align with their objectives and tailor their approach accordingly.
\section{Conclusion and Broader Impact}
We have introduced TROPE, a training-free method for zero-shot captioning that enhances base captions by adding semantic part proposals to key object instances. This approach has demonstrated state-of-the-art performance in fine-grained zero-shot image captioning (IC), consistently improving captions across all tested models, metrics, and datasets. Given the foundational role of IC in a variety of Vision-Language tasks, TROPE holds potential for enhancing fine-grained performance in applications such as text-to-image generation, text-to-image retrieval, and image-to-text retrieval. Future work could also focus on extending the principles underlying TROPE to other modalities, such as audio or video. This would involve adapting TROPE to work with relevant pre-trained models tailored to these modalities, potentially opening new avenues for multimodal integration and captioning enhancements. 
\section{Limitations}
Because TROPE relies on inferences from pre-trained IC models, domains where these pre-trained models have little familiarity with the constituent objects, parts, and terminology like medical imagery are likely to yield very poor zero-shot IC results. These limitations are also applicable to the other training-free baselines presented in this work and could possibly be mitigated with domain-specific human annotation as explored in few-shot or text-based training methods. For high-risk applications, practitioners should examine the overlap between the utilized detectors vocabulary and objects commonly present in the target domain. In such applications, including more than a single semantic part proposal should only be considered if this overlap is high, which reduces the risk of decreasing base caption precision.  
\section{Ethics Statement}
Bias in pre-trained IC models \citep{capbias2018,bias2019} is a concerning challenge for researchers that can potentially impact gender and racial inclusion \citep{women2018}. Zero-shot settings are especially susceptible to carrying over bias from the training dataset since no test set data is available. The use of object detector-based primitives in zero-shot settings could be a promising avenue for mitigating bias in a concise and explainable manner. TROPE has the potential to improve the diversity of generated captions and models trained using those captions. This in turn could improve the inclusion of different genders and races.
\section{Acknowledgements}
TROPE is supported by NSF Robust Intelligence program grants \#1750082 and \#2038666. The authors acknowledge technical access (through ASU-OpenAI collaboration) and support from ASU Enterprise Technology.
The views and opinions of the authors expressed herein do not necessarily state or reflect those of the funding agencies and employers. 
\bibliography{custom}

\end{document}